\documentclass[a4paper,UKenglish,cleveref, autoref, thm-restate]{arxiv}

\usepackage[disable]{todonotes}

\usepackage{amsfonts}
\usepackage{listings}

\usepackage{comment}
\usepackage[normalem]{ulem}
\usepackage{color,soul}
\usepackage{soul}
\usepackage{url}
\usepackage[utf8]{inputenc}
\usepackage{caption}
\usepackage{graphicx}
\usepackage{amsmath}
\usepackage{amsthm}
\usepackage{booktabs}
\usepackage{algorithmic}
\newcommand{\ltlf}{LTL$_f$}
\newcommand{\loss}{$\mathcal{L}$}
\newcommand{\dloss}{$d\mathcal{L}$}
\newcommand{\comp}{\texttt{comp}}
\newcommand{\constr}{\texttt{constraint}}

\title{Constrained Training of Neural Networks via Theorem Proving}

\author{Mark {Chevallier}}{School of Informatics, University of Edinburgh, UK}{m.chevallier@sms.ed.ac.uk}{https://orcid.org/0000-0001-5307-7018}{}
\author{Matthew {Whyte}}{School of Informatics, University of Edinburgh, UK}{m.j.whyte@sms.ed.ac.uk}{}{}
\author{Jacques D. {Fleuriot}}{School of Informatics, University of Edinburgh, UK}{jdf@ed.ac.uk}{https://orcid.org/0000-0002-6867-9836}{}

\authorrunning{M. Chevallier, M. Whyte and J.\,D. Fleuriot}

\Copyright{Mark Chevallier, Matthew Whyte and Jacques D. Fleuriot}



\nolinenumbers
\hideCopyright{}

\begin{document}

\maketitle

\begin{abstract}
We introduce a theorem proving approach to the specification and generation of temporal logical constraints for training neural networks. We formalise a deep embedding of linear temporal logic over finite traces (\ltlf) and an associated evaluation function \texttt{eval} characterising its semantics within the higher-order logic of the Isabelle theorem prover. We then proceed to formalise a loss function \loss{} that we formally prove to be sound, and differentiable to a function \dloss.~We subsequently use Isabelle's automatic code generation mechanism to produce OCaml versions of \ltlf, \loss{} and \dloss{} that we integrate with PyTorch via OCaml bindings for Python. We show that, when used for training in an existing deep learning framework for dynamic movement, our approach produces expected results for common movement specification patterns such as obstacle avoidance and patrolling. The distinctive benefit of our approach is the fully rigorous method for constrained training, eliminating many of the risks inherent to ad-hoc implementations of logical aspects directly in an ``unsafe" programming language such as Python. 
\end{abstract}

\section{Introduction}

The usefulness and general applicability of neural networks is well understood, but there are several ways in which their behaviour could be improved. In particular, it can be hard to understand the behaviour of a neural network, as they take a black box approach in which it is usually difficult to establish the reason why a given result was achieved. 

Additionally, a lot of training is often needed before a neural network behaves as desired, and uncertainty in the behaviour of a neural network can make it entirely unsuited to certain safety-critical tasks such as robot movement.

If we can use some form of logical specifications or constraints as part of the training of a neural network then this can have benefits such as:
\begin{itemize}
	\item The network's output can be interpreted in light of the specifications passed to it. A logical specification is clear and well-defined, and if it is violated by the output, this can provide a way of assessing the network.
	\item The volume of data required to train the network to ensure that it does not breach the constraints could be reduced e.g. where a specification may be simple to express but need many sets of training data for a neural network to learn through imitation \cite{7487517}.
\end{itemize}
Of particular relevance to the current work is the approach by Innes and Ramamoorthy \cite{innes2020elaborating_rss}, which builds on work by Fischer \cite{fischer2019dl2} and aims to improve the training speed of learning robotic movement by mimicking a demonstrator. The suggestion is that given some logical criterion specified in linear temporal logic (e.g.\ ``don't tip the cup until you are above the bowl''), the network learning the movements will need less demonstrator data to have confidence that it will not breach the rule, even on unseen inputs. We extend this work by taking a fully-rigorous, theorem-proving based approach to the logical underpinnings of the loss function and its derivative. 

In particular, by formalising a deep embedding of linear temporal logic over finite traces (\ltlf) and its semantics in the higher-order logic (HOL) of the proof-assistant Isabelle, we formulate a loss function \loss{} that measures whether a statement is satisfied. Moreover, we formally prove that \loss{} is differentiable, with an explicit derivative \dloss{} that can be used as part of the gradient-descent minimisation of the loss.

We use Isabelle to automatically generate OCaml code from our provably correct formal specifications, and integrate it into a PyTorch neural network via a library that provides OCaml bindings for Python. We use both the loss function and its derivative in combination with the usual training process so that the neural network can learn from both training data and our specified constraints. The loss function we generate could in general be applied to any neural network with a notion of time-sequences -- for the purposes of our paper, we demonstrate this using a specific network that predicts paths of motion.

More specifically, we demonstrate experimentally that the neural network learns constrained behaviour when given a wide variety of logical constraints in \ltlf{}. Because our code for the logical implementations of the specifications, the loss function and its derivative was generated automatically by Isabelle, we argue that our approach provides enhanced guarantees about the correctness of the whole pipeline.

The paper is organised as follows: we briefly review some related work next and provide some background to our work. Then in Section \ref{sec:formal}, we discuss \ltlf{} and its formalisation. In Section \ref{sec:ltl} we build our loss function and its derivative, showing both are correct. Section \ref{sec:codegen} discusses how we generated and implemented the OCaml code in PyTorch. In Section \ref{sec:exper} we demonstrate that our code works as expected. We conclude with some thoughts about the work's potential impact and future direction in Section \ref{sec:disc}.

\subsection{Related work}

There has been some work aimed at unifying propositional logical constraints with neural networks that have probabilistic outputs \cite{pmlr-v80-xu18h}. Hu et al.\ trained a network to follow a rules-based ``teacher'' \cite{hu2016harnessing} while work by Li et al.\ incorporates first-order logic rules directly in the network design, with the aim of guiding training and prediction \cite{li2019augmentingnnfol}. Other authors modify the loss function based on the satisfaction of logical constraints to train a neural network \cite{BADREDDINE2022103649,fischer2019dl2,10.1007/978-3-030-46147-8_17}. On the reinforcement learning front, there has been recent work, e.g.\ on specifying policy learning via LTL instructions \cite{kuo2020encoding,pmlr-v139-vaezipoor21a}. None of the above approaches involves any theorem proving like the current work.

A distinct but related strand of work centres around the formal verification of neural networks. This involves using SMT solvers to formally verify properties of neural networks \cite{ehlers2017formal,huang2017safety,katz2017reluplex,scheibler2015towards}. These SMT solvers typically verify simple propositional logic constraints over boolean variables, and in any case are not used to train the network but to check it. This is distinct from our objective, where we are formally verifying the logical system used to train a neural network and automatically generating code for the actual training.

\subsubsection{Comparison with Innes and Ramamoorthy's work}

As mentioned already, our work is motivated by that of Innes and Ramamoorthy \cite{innes2020elaborating_rss}.~However, our examination of their approach, and in particular, the fidelity of the code that implements their experiments with the paper's descriptions, uncovered several weaknesses that, we believe, support the need for a more formal neurosymbolic pipeline. 

The Python code for their logical formulation does not always match that given in their paper: the ``$\neq$'' comparison, for example, is given in the paper as an indicator function, but in the Python script, it is defined as $a\neq b \!\!\iff\!\! (a < b) \vee (b < a)$. Moreover, although presented in their paper, they do not encode a component of the loss function for the LTL \texttt{Until} operator, meaning they were not able to cover any tests involving it -- something which we can do in our work (see Section \ref{ssec:ltlf}). 

There are also some aspects of LTL in Innes and Ramamoorthy's paper that remain implicit and are not discussed -- notably its use of LTL is over finite traces, which has a distinct semantics from the more usual formulation of LTL over infinite traces.

The primary advantage of our work in comparison to theirs resides in our fully rigorous specifications and proofs, with the  training constraints guaranteed via systematic code generation from our specifications. This guarantee means we know that the code generated will have the properties that were established for it during the theorem proving stage of our process.

\subsection{Isabelle}
\label{ssec:isabelle}

We briefly review a few aspects of Isabelle \cite{nipkow2002isabelle} that are relevant to this work. Mathematical theories written in Isabelle are a collection of formal definitions of various kinds (algebraic objects, types, functions, etc.), and associated theorems about their properties. Proofs can be written in a pen-and-paper like, structured proof language \cite{wenzel1999isar}. 

When defining a constant or function, it must be assigned a type (``\texttt{T::$\tau$}'' states that \texttt{T} is of type \texttt{$\tau$}). Function types are stated using ``\texttt{$\Rightarrow$}''. We provide some quick illustrative examples:
\begin{enumerate}\item \texttt{state :: "int $\Rightarrow$ real"} tells us that a \texttt{state} is a function from the \texttt{int} type, the integers, to the real numbers. Note also that functions are usually curried. \item \texttt{path :: "state list"} tells us that a \texttt{path} is a finite list of such states i.e. a list of functions.\end{enumerate}
\noindent Note that in Isabelle, a list can be written as \texttt{s\#ss} denoting the element \texttt{s} being consed onto the (possibly empty) list \texttt{ss} and that the lambda abstraction, or anonymous function, is denoted by \texttt{$\lambda$x.\ M} so e.g. \texttt{$\lambda$x.\ x$^2$} denotes the function that squares its argument \texttt{x}. 

It is possible to generate computable code from the formal specification of functions in Isabelle into Standard ML (PolyML), OCaml, Haskell and Scala \cite{haftmann_codegen}. This mechanism provides a rigorous link between Isabelle concepts, e.g.\ our \loss{} and \dloss{} functions, and their automatically-generated counterparts, whose computational behaviour (modulo implementational details such as translating reals to floats) can then be expected to respect the properties that were formally proven as theorems in Isabelle. This is a vital component of our pipeline and will be discussed further in Section \ref{sec:codegen}.

\section{Formalising linear temporal logic}
\label{sec:formal}
As already mentioned, We mechanise linear temporal logic over finite paths in Isabelle and use it to formulate a loss function for training a neural network under rigorously specified logical constraints. We review some of the salient aspects of our formalisation next.

\subsection{States and paths}
\label{ssec:statepath}
\begin{figure}
\begin{center}
    \includegraphics[width=0.4\textwidth]{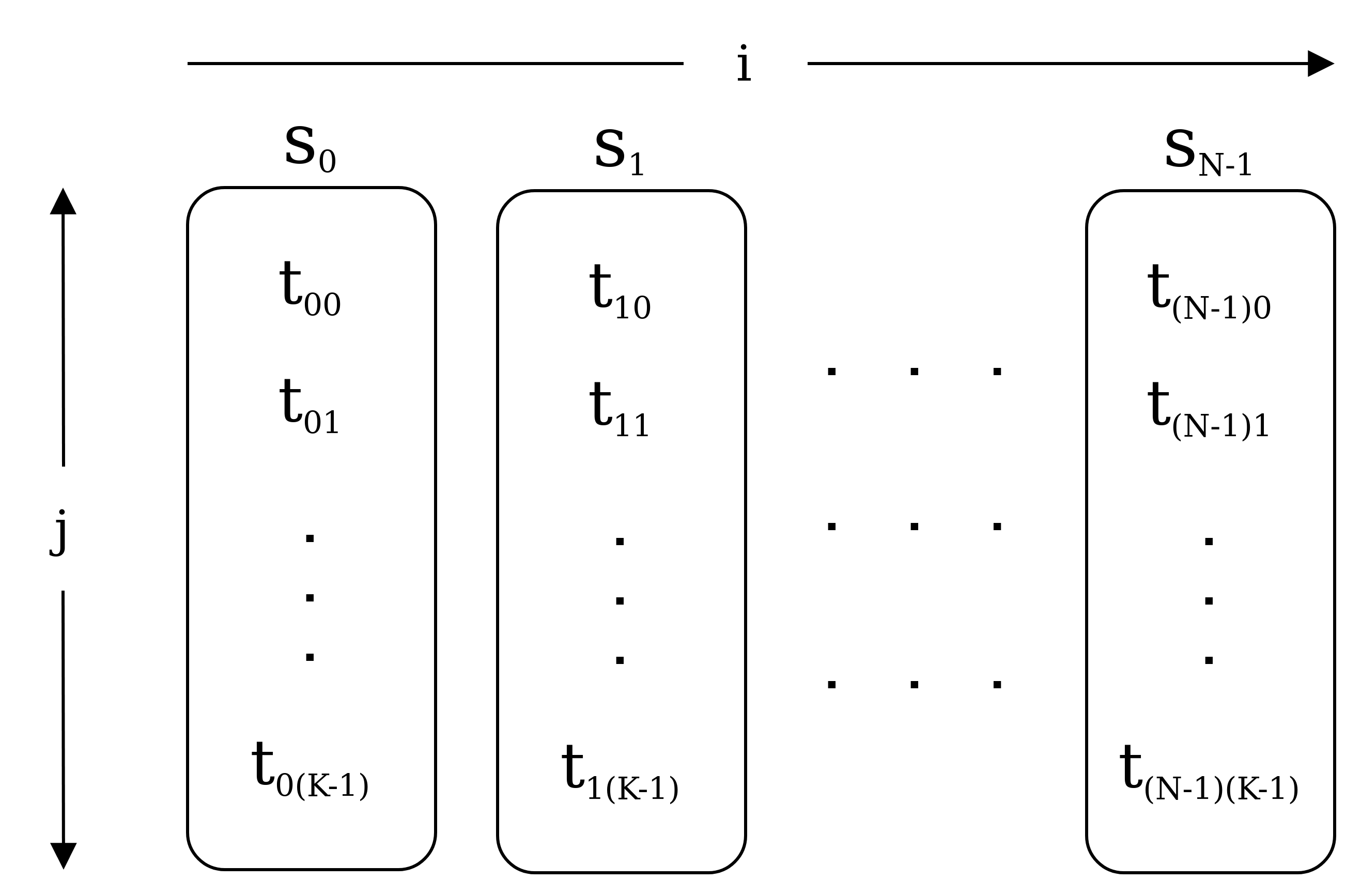}
    \caption{A path of states with values: time proceeds from left to right and j indexes the values each state can measure}
    \label{fig:pathstate}
\end{center}    
\end{figure}

Recall the definition of \texttt{state} from the previous section. In the context of our work, the state function at a particular time-step $i$ tracks several values in a learning problem that might be compared: a constant, or some measurement that a neural network uses in its training e.g.\ how far a robotic hand is from a barrier at a given moment in time. The integers passed to a state function are simply the indices to those values of interest. In almost all cases, a state function only needs to represent some finite number $K$ of values. By only considering indices $j$ over this range, the state function for a given time-step can be considered as a vector, as can be seen in Fig.\ \ref{fig:pathstate}.\footnote{Strictly speaking, we could have used natural numbers as our indices. However, we chose integers because OCaml, the target language for our specifications, has no native natural number type.}

With each state function encoding information about a system at a specific time-step, a \texttt{path} of length $N$ encodes the evolution of a system over $N$ time-steps. In the case of Fig.\ \ref{fig:pathstate}, increasing $i$ represents the forward temporal progression of the system. It should also be clear from this figure that the values encoded by a \texttt{path} of \texttt{state}s can equivalently be represented in matrix form, which will be of vital importance to our PyTorch integration (see Section \ref{ssec:pytorchint}).

\subsection{LTL and finite traces of states}
\label{ssec:ltlf}

Our formalisation uses a variant of LTL \cite{pnueli1977} known as \ltlf{} \cite{de2013linear}, which is interpreted over finite traces (of states) and is often viewed as a more natural choice for applications in AI, e.g.\ planning \cite{baier2006}, where processes are usually of finite duration. Note that in what follows, any reference to LTL means \ltlf{} unless otherwise stated.

As we are using LTL to evaluate conditions during the training of a neural network, we take comparisons of real-valued terms (each term corresponding to measurements in the environment or constants), namely $t < t', t \leq t', t = t', t \neq t'$ as our atomic propositions $\rho$ \cite{fischer2019dl2}. Thus we divide constraints into these comparisons (\comp)  and those constraints (\constr) arising from LTL's operators (see further down for their Isabelle implementations).

The \ltlf{} formulae are thus: $\rho$, $\rho_1 \wedge \rho_2$, $\rho_1 \vee \rho_2$, $\mathcal{N}\rho$ (Next), $\square \rho$ (Always), $\Diamond \rho$ (Eventually), $\rho_1 \mathcal{U} \rho_2$ (Weak Until), $\rho_1 \mathcal{R} \rho_2$ (Strong Release). In \ltlf{} these have the usual semantics of LTL, except at the end of a path. In particular, for a sequence of length $i$, $\neg(\mathcal{N} \rho)$ holds at the final step $i$ \cite{de2013linear}. As a consequence, when dealing with a finite time-sequence of length $i$, it is important to recognise that $\neg(\mathcal{N} \rho) \!\!\iff\!\! \mathcal{N} (\neg\rho)$ is not true. $\neg(\mathcal{N} \rho)$ is true at the last step in a time-sequence and $\mathcal{N} (\neg\rho)$ is false at that step. We refer the reader to Appendix \ref{app:ltlf} for some more details about \ltlf{}.

We formalise negation via a \texttt{Not} function which, given any constraint, returns a constraint that is provably logically equivalent to its negation. This reduces the number of primitive operators that one needs to specify for \ltlf{}, thereby simplifying reasoning about its properties. 

We proceed by defining Isabelle datatypes \comp{} and \constr{} for the comparison and constraints respectively. This approach to specifying the language in (higher-order) logic is known as a deep embedding \cite{boulton1992experience}. Doing so will enable us to prove that the loss function is sound and, importantly, generate fully self-contained code for the specification language that will be used as part of the training of our neural net. Note also our formulation of \texttt{state} and \texttt{path} which, as per the previous section, creates a finite time-sequence over which \ltlf{} can be evaluated.

\begin{lstlisting}[basicstyle=\footnotesize\ttfamily, mathescape = true,caption={Datatypes for our \ltlf{ }implementation},captionpos=b]{}
type_synonym state = "int $\Rightarrow$ real"
type_synonym path = "state list"

datatype comp = Less int int | Lequal int int | Equal int int 
    | Nequal int int

datatype constraint = Comp comp | And constraint constraint 
    | Or constraint constraint | Next constraint | Always constraint 
    | Eventually constraint | Until constraint constraint 
    | Release constraint constraint
\end{lstlisting}

Next, we formalise our \texttt{eval} function, which characterises the semantics of \ltlf{}  by recursively evaluating the truth-value of a constraint over a path. We give an extract of the Isabelle definition (see Appendix \ref{app:listings} for the full definition):

\begin{lstlisting}[basicstyle=\footnotesize\ttfamily, mathescape = true,caption=An extract of the {\ltlf{ }evaluation function},captionpos=b]{}
function eval :: "constraint $\Rightarrow$ path $\Rightarrow$ bool" where
  "eval c [] = False"
| "eval (Comp (Lequal v1 v2)) (s # ss) = ((s v1) $\leq$ (s v2))"
  ...
| "eval (Always c) (s # ss) = ((eval c (s # ss)) 
    $\wedge$ (if ss = [] then True else (eval (Always c) ss)))"
 ...
\end{lstlisting}

Given the complexity of \ltlf{} compared to propositional logic, we also prove a number of \ltlf{} equivalences, which confirms that our \texttt{eval} behaves as expected. For example, we show (amongst other examples) that, $\square(\square \rho) = \square \rho$ and that $\mathcal{N}(\rho_1 \vee \rho_2) = (\mathcal{N}\rho_1) \vee (\mathcal{N}\rho_2)$.
\section{A LTL-based loss function and its derivative}
\label{sec:ltl}

The loss function $\mathcal{L}$ -- which takes a constraint $\rho$, a path $P$ and a relaxation factor $\gamma$, and returns a real value -- needs to satisfy several important properties:
\begin{enumerate}
	\item $\mathcal{L}(\rho, P, \gamma) \geq 0$; \label{prop1}
	\item $\mathcal{L}$ is differentiable with respect to any of the terms that the constraint compares; \label{prop2}
	\item (Soundness) $\lim_{\gamma\to0}\mathcal{L}(\rho,P, \gamma) = 0 \!\!\iff\!\! \rho(P)$, where $\rho(P)$ is the truth value of $\rho$ on $P$. \label{prop3}
\end{enumerate}

\subsection{Soft functions and their derivatives}
\label{ssec:soft}

When formalising $\mathcal{L}$, in order for it to be differentiable, it is necessary to use {\it soft} versions of various functions. Thus, based on the work by Cuturi and  Blondel \cite{cuturi2017soft}, we define binary softmax and softmin functions, $\max_\gamma$ and $\min_\gamma$, respectively, as: 
\begin{align*}
\max_\gamma\;  a_1 \; a_2  &= 
\begin{cases}
\max a_1 \; a_2 &\gamma \leq 0\\ 
\gamma \log (e^{a_1/\gamma} + e^{a_2/\gamma}) &\text{otherwise}
\end{cases}
\\
\min_\gamma\;  a_1 \; a_2  &=
\begin{cases}
\min a_1 \; a_2 &\gamma \leq 0\\ 
-\gamma \log (e^{-a_1/\gamma} + e^{-a_2/\gamma}) &\text{otherwise}
\end{cases}
\end{align*}

Each of these soft functions takes an additional parameter $\gamma$. The intention behind these is that as $\gamma\to0$, $\max_\gamma \to \max$, $\min_\gamma\to\min$, and that they are differentiable for $\gamma>0$. In Isabelle, these are easily formalised and proven as correct -- we give \texttt{Max\_gamma} as an example (see Appendix \ref{app:listings} for \texttt{Min\_gamma}):

\begin{lstlisting}[basicstyle=\footnotesize\ttfamily, mathescape = true,caption={The soft gamma Max function and its correctness theorem},captionpos=b]{}
fun Max_gamma :: "real $\Rightarrow$ real $\Rightarrow$ real $\Rightarrow$ real" where
  "Max_gamma $\gamma$ a b = (if $\gamma$ $\leq$ 0 then max a b 
     else $\gamma$ * ln (exp (a/$\gamma$) + exp (b/$\gamma$)))"

lemma Max_gamma_lim: "($\lambda$$\gamma$. Max_gamma $\gamma$ a b) $-$0$\rightarrow$ max a b"
\end{lstlisting}
\noindent where \texttt{$(\lambda x. f \, x)$ $-$0$\rightarrow$ L} denotes 
$\lim_{x\to 0} f(x) = L$ in Isabelle \cite{fleuriot2000}. 

For \texttt{Max\_gamma}, the derivatives with respect to $a$ and $b$ are built separately before being combined to give the \texttt{dMax\_gamma\_ds} function (as defined in Appendix \ref{app:soft}). We show that this is the expected derivative using Isabelle's  \texttt{has\_real\_derivative} relation by proving the following theorem (for $\gamma > 0$):

\begin{lstlisting}[basicstyle=\footnotesize\ttfamily, mathescape = true,caption={The derivative's correctness stated in Isabelle},captionpos=b]{}
theorem dMax_gamma_chain:
  assumes "$\gamma$ > 0" and "(f has_real_derivative Df) (at x)" 
    and "(g has_real_derivative Dg) (at x)"
 shows "(($\lambda$y. Max_gamma $\gamma$ (f y) (g y)) has_real_derivative 
        (dMax_gamma_ds $\gamma$ (f x) Df (g x) Dg)) (at x)"    
\end{lstlisting}

Likewise, we use other soft functions to capture losses from the \texttt{Nequal} comparison, again using $\gamma$ as a parameter. We note here an important distinction between our approach and that of Fischer \cite{fischer2019dl2} and Innes and Ramamoorthy \cite{innes2020elaborating_rss}: in their respective work, the \texttt{Nequal} and \texttt{Lequal} comparisons were not defined using soft-functions and were not differentiable. They relied on the implicit auto-differentiation machinery of PyTorch to handle these for backpropagation purposes. In our case, though, we provide an explicit derivative of the loss function to PyTorch thereby giving us guarantees that backpropagation is using the desired function for gradient descent (see Sections \ref{sec:dloss} and \ref{ssec:pytorchint}). We therefore require the loss functions for all our comparisons to be provably differentiable. The interested reader is referred to Appendix \ref{app:soft} for more detail.

\subsection{Formalising the loss function}
\label{ssec:loss}

We proceed to define $\mathcal{L}$ recursively over the constraint, given a path and a relaxation factor $\gamma$. We give an extract of the definition:
\newpage 
\begin{lstlisting}[basicstyle=\footnotesize\ttfamily, mathescape = true,caption={An extract of the \loss{ }function, see Appendix \ref{app:listings} for the full definition},captionpos=b]{}
  function L :: "constraint $\Rightarrow$ path $\Rightarrow$ real $\Rightarrow$ real" where
    "L c [] $\gamma$ = 1"
  | "L (Comp (Lequal v1 v2)) (s # ss) $\gamma$ = Max_gamma $\gamma$ (s v1 - s v2) 0"
  | "L (Comp (Nequal v1 v2)) (s # ss) $\gamma$ = Nequal_gamma $\gamma$ (s v1) (s v2)"
   ...
  | "L (Always c) (s # ss) $\gamma$ = Max_gamma $\gamma$ (L c (s # ss) $\gamma$) 
       (if ss = [] then 0 else (L (Always c) ss) $\gamma$)"
   ...
  | "L (Until c1 c2) (s # ss) $\gamma$ = Min_gamma $\gamma$  (L c2 (s # ss) $\gamma$) 
         (Max_gamma $\gamma$ (L c1 (s # ss) $\gamma$) 
           (if ss = [] then 0 else (L (Until c1 c2) ss $\gamma$)))"
   ...
\end{lstlisting}

In the above formulation, the defining equation for the \texttt{Lequal} shows that if the first state-value is smaller or equal to the first, \loss{} produces 0, equivalent to logical truth. This works in a similar way to a soft rectification function -- its limit as $\gamma$ tends to zero is identical and proven in Isabelle. As another remark, our comparison operators for \loss{} are defined in terms of $\leq$ and $\neq$, with the other two comparisons, $<$ and $=$ defined using them.

For all the LTL operators, the \loss{} function, in common with our \texttt{eval} function, recurses over the constraint from the outside in, and recurses down the path as required for temporal operators. Innes and Ramamoorthy do not use a recursive definition \cite{innes2020elaborating_rss}, which though fine for relatively simple LTL operators such as \texttt{Always}, leads to a much more involved formulation for more complicated ones such as \texttt{Until}.

In particular, our recursive definition of \loss{} against the \texttt{Until} operator is significantly simpler than that given in their paper because in our case \texttt{Until} is logically equivalent to:
$$\rho_1\mathcal{U}\rho_2(i)=(\rho_2(i)\vee(\rho_1(i)\wedge(\rho_1\mathcal{U}\rho_2(i+1))$$ 
\noindent where the $\rho_1\mathcal{U}\rho_2(i+1)$ clause is assumed to be true at the path's termination. 

Note also that when we are evaluating over an empty path, $\mathcal{L}$ takes value 1, which is equivalent to any constraint on the empty path evaluating as false. This means that our \texttt{Next} constraint matches our expectation (as discussed in Section \ref{ssec:ltlf}) at the end of a finite trace. However, this understanding of how \loss{} treats the empty path means we must specify a slightly different behaviour for how \loss{} handles finite paths for the \texttt{Always} and \texttt{Weak Until} constraints. As \loss{} recurses down the path for these two, if it reaches the end of the path, it should return a 0 value (equivalent to true) if every state it has recursed through meets the specified constraints.

Once we have formulated $\mathcal{L}$ via a series of lemmas and an inductive proof on the \texttt{constraint} datatype, we show that it has the expected property with respect to the \texttt{eval} function. The proof is not straightforward and depends on being able to show continuity of our various soft gamma functions when $\gamma=0$.

\begin{lstlisting}[basicstyle=\footnotesize\ttfamily, mathescape = true,caption={Soundness of \loss{} stated in Isabelle},captionpos=b,label=sound]{}
theorem L_sound: "(($\lambda$$\gamma$. L c ss $\gamma$) $-$0$\rightarrow$ 0) $\longleftrightarrow$ (eval c ss)"
\end{lstlisting}

We have now formally defined an LTL-based, soft loss function $\mathcal{L}$ that, for any constraint and finite trace, tends to 0 as its gamma parameter tends to 0, if and only if the constraint is satisfied over that trace.

\subsection{Derivative of the loss function}
\label{sec:dloss}

We next construct a derivative \dloss{} for the \loss{} function, to be used for gradient-based methods in PyTorch (see Section \ref{ssec:pytorchint}). The derivative must be defined with respect to each time-step $i$ and state-value index $j$ at that time-step along the finite trace. The full formulation is extensive, so for brevity, we only present an extract here:

\begin{lstlisting}[basicstyle=\footnotesize\ttfamily, mathescape = true,caption={An extract of the \dloss{ }function, see Appendix \ref{app:listings} for the full definition},captionpos=b]{}
function dL :: "constraint $\Rightarrow$ path $\Rightarrow$ real $\Rightarrow$ int $\Rightarrow$ int $\Rightarrow$ real" where
  "dL c [] $\gamma$ i j = 0"
   ...
| "dL (Comp (Nequal v1 v2)) (s # ss) $\gamma$ i j = (if (i $\neq$ 0) then 0 
   else (dNequal_gamma_ds $\gamma$ (s v1) (if j = v1 then 1 else 0) (s v2) 
     (if j = v2 then 1 else 0)))"
   ...
| "dL (Until c1 c2) (s # ss) $\gamma$ i j = dMin_gamma_ds $\gamma$ (L c2 (s # ss) $\gamma$)
    (dL c2 (s # ss) $\gamma$ i j) (Max_gamma $\gamma$ (L c1 (s # ss) $\gamma$) 
    (if (ss = []) then 0 else (L (Until c1 c2) ss $\gamma$)))
    (dMax_gamma_ds $\gamma$ (L c1 (s # ss) $\gamma$) (dL c1 (s # ss) $\gamma$ i j)
    (if (ss = []) then 0 else (L (Until c1 c2) ss $\gamma$))
      (if (ss = []) then 0 else (dL (Until c1 c2) ss $\gamma$ (i-1) j)))"
   ...
\end{lstlisting}

The \dloss{} function formalisation is structured similarly to our formalisation of the \loss{} function, defined recursively over the components of the \ltlf{ }constraint passed to it and essentially follows from repeated applications of the chain rule. In defining it, we make extensive use of the derivatives of the soft-functions we described in Section \ref{ssec:soft} (and in Appendix \ref{app:soft}).

\subsubsection{Correctness of \dloss{}} 

While the formulation of \dloss{} may look intricate, we formally prove that it is indeed the correct derivative for our loss function and thus guarantee that when used for backpropagation it will achieve the desired results.

In order to facilitate this, we must turn a \texttt{path} into a function of a single variable representing a single state-value at a specific time-step on the path. We do so by formalising the state-update function \texttt{update\_state} and the recursive function \texttt{update\_path}, which allow us to specify the value at a particular $i$ and $j$:

\begin{lstlisting}[basicstyle=\footnotesize\ttfamily, mathescape = true,caption={The update functions},captionpos=b]{}
fun update_state :: "state $\Rightarrow$ int $\Rightarrow$ real $\Rightarrow$ state" where
  "update_state s j x = ($\lambda$n. if n=j then x else (s n))"

primrec update_path :: "path $\Rightarrow$ int $\Rightarrow$ int $\Rightarrow$ real $\Rightarrow$ path" where
  "update_path [] i j = ($\lambda$x. [])"
  | "update_path (s#ss) i j = (if i = 0 then 
    ($\lambda$x. (update_state s j x) # ss) else (if i < 0 then ($\lambda$x. (s#ss))
    else ($\lambda$x. s # (update_path ss (i-1) j x))))"
\end{lstlisting}

\noindent Using this mechanism, we prove that \dloss{} is indeed the derivative of the \loss{} function, with respect to the value at any $i$ and $j$. In Isabelle, the theorem formalising this is as follows:

\begin{lstlisting}[basicstyle=\footnotesize\ttfamily, mathescape = true,caption={Correctness of \dloss{} stated in Isabelle},captionpos=b]{}
theorem L_has_derivative: 
  assumes "$\gamma$ > 0"
  shows "(($\lambda$y. L c (update_path pth i j y) $\gamma$) has_field_derivative 
     (dL c (update_path pth i j x) $\gamma$ i j)) (at x)"
\end{lstlisting}

\section{A PyTorch-compatible LTL loss function}
\label{sec:codegen}

With $\mathcal{L}$ and $d\mathcal{L}$ formalised in Isabelle, what remains is to integrate them into the PyTorch environment. Unfortunately, there does not exist a mechanism for generating Isabelle functions as Python code. Instead, we choose to generate intermediate representations of $\mathcal{L}$ and $d\mathcal{L}$ in OCaml since our recursive Isabelle functions can be straightforwardly translated to type-safe OCaml ones and, moreover, there exists a Python library that can be used to call OCaml functions from within Python code \cite{ocamlpython}. 

\subsection{OCaml code generation}
\label{ssec:ocamlgen}

In order to produce computable code, we need to map the real numbers of Isabelle to floating points. As this is an approximation, it naturally has some scope for machine arithmetic errors, although the code generated for the various functions is fully faithful to their definitions in Isabelle. So, assuming that the floating point computations are well-behaved, we expect the OCaml functions to satisfy the properties that were proven for their Isabelle counterparts (e.g.\ that the OCaml-generated \dloss{} is the derivative of the OCaml-generated \loss{}). 

We make use of code generation machinery of Isabelle, which provides code printing instructions for translating between real numbers in Isabelle to floating point numbers in OCaml \cite{codereal} to generate an OCaml module \texttt{LTL\_Loss}.

\subsection{PyTorch integration}
\label{ssec:pytorchint}

Python bindings for the OCaml definitions of \loss{} and \dloss{} are incorporated into a PyTorch \texttt{autograd.Function} object through the \texttt{forward} and \texttt{backward} methods, respectively. These methods are required for the loss function to form part of a computational graph in PyTorch and enable training based on gradient descent.~Each object is parameterised by an LTL constraint, represented as an OCaml expression; constants for comparison; and a value for $\gamma$. Consequently, our \texttt{LTL\_Loss} module, implemented as a subclass of \texttt{autograd.Function}, is functionally identical to a differentiable PyTorch operation on tensors. Paths are represented as tensors, with each row containing the values related to a particular state. This is a transpose of the arrangement depicted in Fig.~\ref{fig:pathstate}.

Significantly, we know exactly how \texttt{LTL\_Loss} should behave when computing gradients with \texttt{autograd}, as it is solely characterised by its formalisation in Isabelle. For a more detailed explanation on the construction of the class and the mapping of a path to a PyTorch tensor, the reader is referred to Appendix \ref{app:pytorchint}.

\section{Experiments}
\label{sec:exper}

With our Isabelle-formalised loss function $\mathcal{L}$ and its explicit derivative $d\mathcal{L}$ fully available as a generic \texttt{autograd} function in PyTorch, we now verify that we can achieve experimental results that include and extend those of Innes and Ramamoorthy \cite{innes2020elaborating_rss}.

We take our experimental models from their work and extend them to show the improvements provided by our approach. Specifically, after replicating some of the main results from their work using our method, we also demonstrate an application of the \texttt{Until} constraint whose loss evaluation was not implemented in their code.

\subsection{Domain setup}

Each of the tests takes place in a 2-dimensional planar environment with Cartesian co-ordinates. The training data follow a spline-shaped curve consisting of $N\!\!=\!\!100$ sequenced points in the plane following the curve with small random perturbations, simulating a demonstrator moving via some trajectory from the origin to some destination in the plane. We train a feed-forward neural network to learn a Dynamic Movement Primitive (DMP) \cite{schaal2005learning} to follow this trajectory.

\subsection{Unconstrained training}

Let $D$ denote the trajectory of the demonstrator along the spline and $Q$ denote the trajectory of the DMP learned by the neural network. Both $Q$ and $D$ are $N \times 2$ matrices. Moreover, let $Q_i$ denote the row vector at index $i$ of $Q$ (likewise for $D_i$ and $D$). The per-sample imitation loss, $L_d$, for this sample pair is given by $L_d(Q,D) = \frac{1}{N}\sum_{i=0}^{N-1}{\|Q_i - D_i\|^2}$.

Intuitively, $L_d$ penalises the learned trajectory for deviating from the demonstrator. For a batch of samples, we compute the average imitation loss. $L_d$ is used as the sole loss in the training of the neural network over 200 epochs using the Adam optimizer \cite{adam} with a learning rate of $10^{-3}$. 

\subsection{Constrained training with \ltlf{}}
\label{ssec:contrain}

First, it is important to distinguish that for a given sample, its trajectory $Q$ is not necessarily the same as the corresponding path $P$ to be reasoned over by LTL. While $Q$ encodes the $x$ and $y$ co-ordinates at each point, $P$ encodes values about the trajectory (which may include intermediate functions of $x$ and $y$) as well as any constants for comparison, that are reasoned over by the LTL constraint.

With this in mind, consider a differentiable function $g$ which maps $Q$ to a path $P$, an LTL constraint $\rho$ which reasons over that path, and a relaxation factor $\gamma>0$. We can incorporate this constraint into the learning process by augmenting the per-sample loss function to be minimised to give $L_{full} = L_d(Q,D) + \eta\mathcal{L}(\rho,g(Q),\gamma)$, where $\eta$ is a positive constant representing the weighting of the constraint violation against the imitation loss. This weighting can be adjusted to reflect the priority in satisfying the constraints relative to following the demonstrator accurately.

We now repeat the same training procedure as for the unconstrained case, but with this new loss function and with $\gamma=0.005$. In PyTorch, the loss \loss{} is implemented by an instantiation of \texttt{LTL\_Loss} (mentioned in Section \ref{ssec:pytorchint}) with $\rho$ and $\gamma$ as arguments.

We lay out 4 different problems. Tests 1, 2, and 4 are similar to those of Innes and Ramamoorthy \cite{innes2020elaborating_rss}, while Test 3 evaluates the \texttt{Until} constraint, which is a non-trivial extension to their available constraints:

\begin{enumerate}
    \item \textbf{Avoid}: The trajectory (always) avoids an open ball of radius $0.1$ around the point $o=(0.4,0.4)$. At each time-step, we compute the Euclidean distance between the trajectory and $o$, which we denote as $p_{do}$. The LTL constraint becomes: $\square (0.1\leq p_{do})$.
    \item \textbf{Patrol}: The trajectory eventually reaches $o_1=(0.2,0.4)$ and $o_2=(0.85,0.6)$ in the plane. With Euclidean distances $p_{do_1},p_{do_2}$, this constraint becomes: $(\Diamond (p_{do_1}\leq 0)) \wedge (\Diamond (p_{do_2}\leq 0))$.
    Note that we do not use the comparison $=$ as both $p_{do_1}$ and $p_{do_2}$ are non-negative by construction and this formulation has a lower computational cost.
    \item \textbf{Until}: The $y$ co-ordinate, $p_y$, of the trajectory cannot exceed $0.4$ until its $x$ co-ordinate, $p_x$, is at least $0.6$: $(p_y\leq 0.4)\text{ }\mathcal{U}\text{ }(0.6\leq p_x)$.
    \item \textbf{Compound}: A more complicated test combining several conditions. The trajectory should avoid an open ball of radius $0.1$ around the point $o_1=(0.5,0.5)$, while eventually touching the point $o_2=(0.7,0.5)$. Further, the $y$ co-ordinate of the trajectory should not exceed $0.8$. With $p_{do_1}$ and $p_{do_2}$ defined in the same way as before, this compound constraint is represented as: $(\square (0.1\leq p_{do_1})) \wedge(\Diamond (p_{do_2} \leq 0))\wedge(\square (p_y\leq0.8))$.
\end{enumerate}

For a more concrete explanation of the role of function $g$, consider the Avoid test. Here, $g$ is defined to act row-wise on a trajectory $Q$, producing new row vectors whose elements are the Euclidean distance $p_{do}$ and the constant $0.1$, as these are the only quantities reasoned over by the LTL constraint.

\subsection{Results}

\begin{table}[t!]
\centering
 \begin{tabular}{|c c c c c|} 
 \hline
 & \multicolumn{2}{c}{Unconstrained} & \multicolumn{2}{c|}{Constrained}\\
 Test & $L_d$ & $\mathcal{L}$ & $L_d$ & $\mathcal{L}$ \\ [0.5ex] 
 \hline\hline
 Avoid & 0.0047 & 0.0789 & 0.0146 & 0.0239 \\ 
 Patrol & 0.0052 & 0.1238 & 0.0312 & 0.0049 \\
 Until & 0.0036 & 0.0970 & 0.0158 & 0.0207 \\
 Compound, $\eta=1$ & 0.0044 & 0.1612 & 0.0318 & 0.0291 \\ 
 Compound, $\eta=4$ & 0.0055 & 0.1611 & 0.0358 & 0.0269 \\ [1ex]
 \hline
 \end{tabular}
 \caption{The losses associated with each test, averaged over 5 iterations}
 \label{table:numresults}
\end{table}

When we run our tests, we first train the neural network ignoring the logical constraint, then we use the latter as part of its loss calculations. The differences between the two sets of results demonstrates the effectiveness of the logical constraint as used in the loss function for the training. The loss figures are shown in Table \ref{table:numresults}.

The losses in the table show that when we do perform unconstrained training, we understandably have high constraint losses. These are substantially improved when we constrain the training, although the imitation losses increase as the trajectory outputted by the neural network no longer follows the training trajectory as closely. 

Different constraints respond with different loss values based on the specific definitions used, so we should be careful when making comparisons between the different constrained tests (save for the two differently weighted compound tests). Even though in most of our tests the constraints are satisfied, the constraint losses are not zero. Given that we are dealing with a soft \loss{} function with a positive $\gamma$ parameter, this is expected. By the soundness theorem (Listing \ref{sound}), the constraint losses would be reduced further with smaller values of $\gamma$.

\begin{figure*}[t!]
    \centering
    \includegraphics[width=0.3\textwidth]{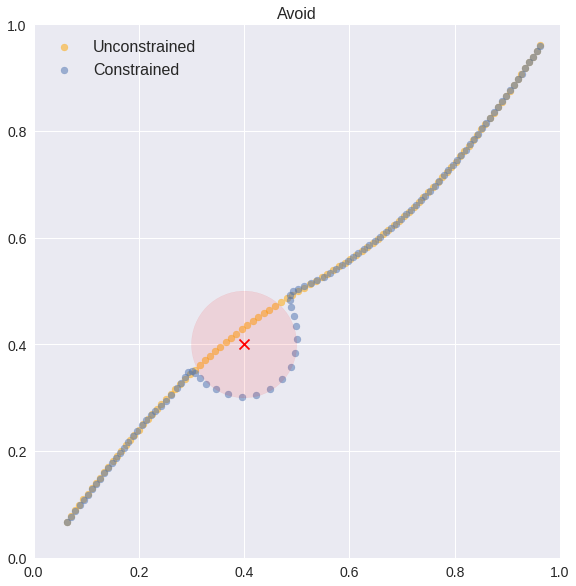}
    \includegraphics[width=0.3\textwidth]{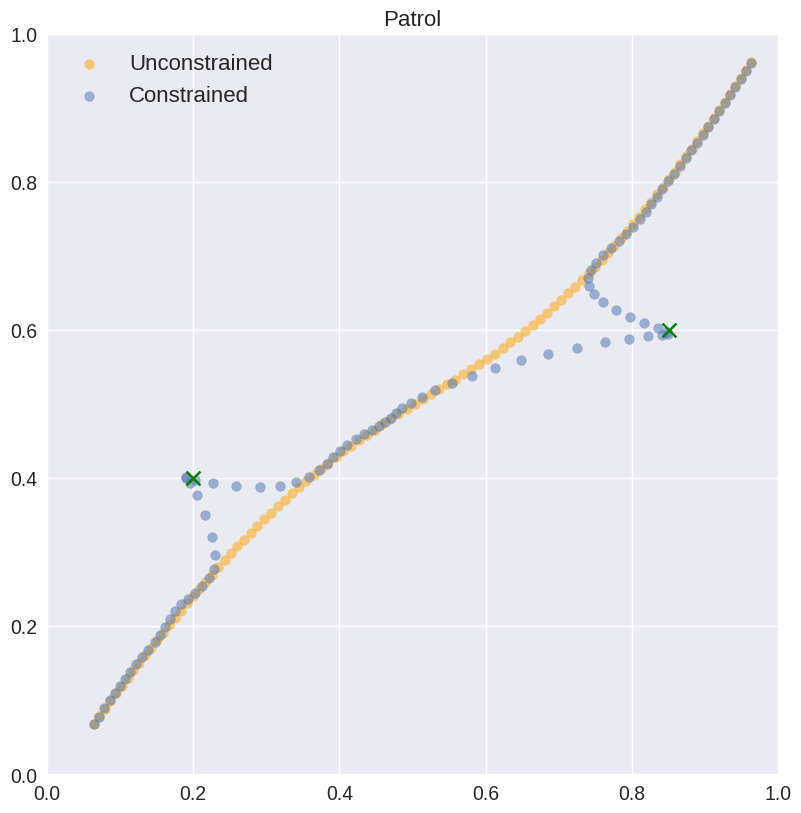}
    \includegraphics[width=0.3\textwidth]{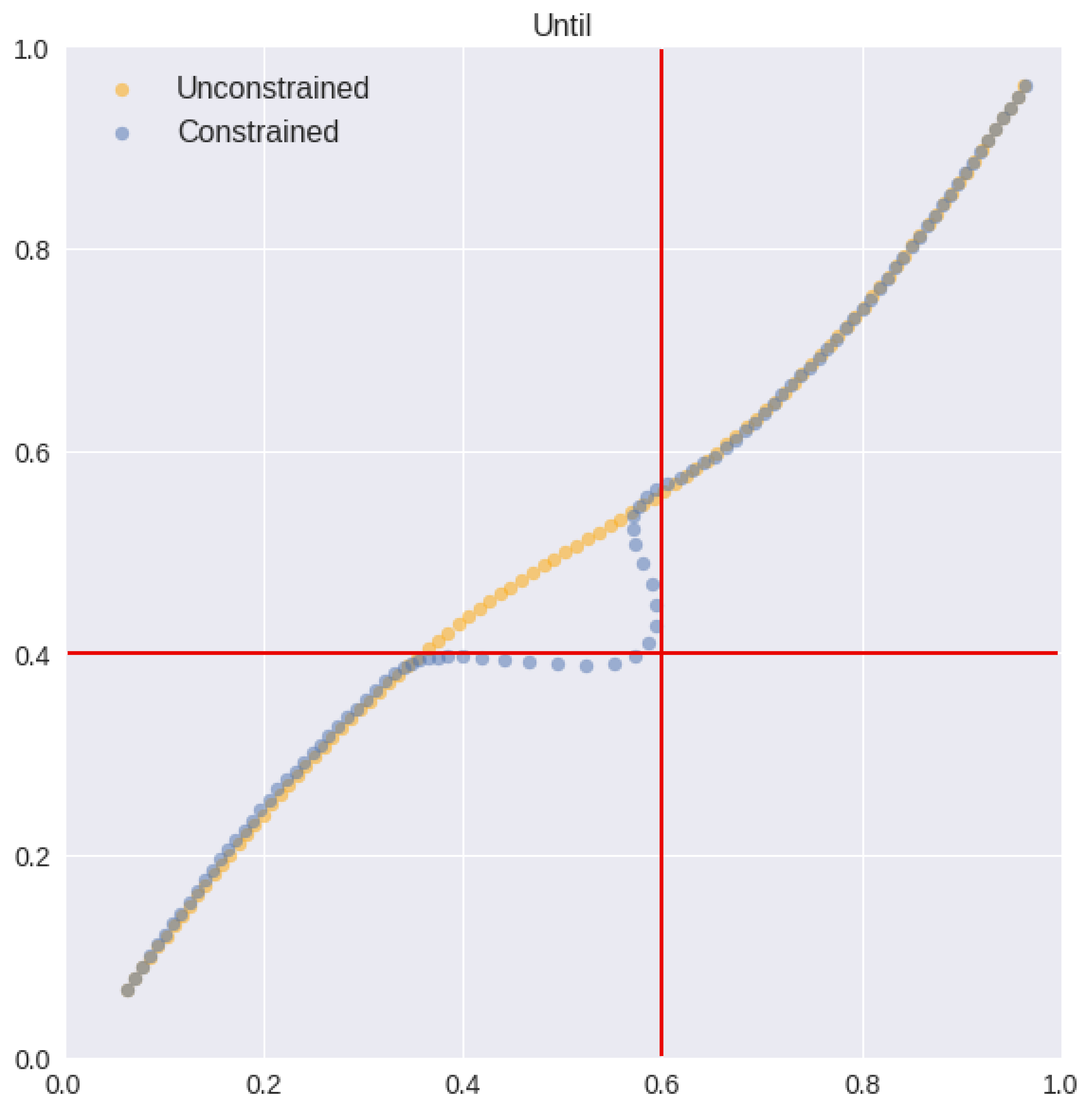}
    \includegraphics[width=0.3\textwidth]{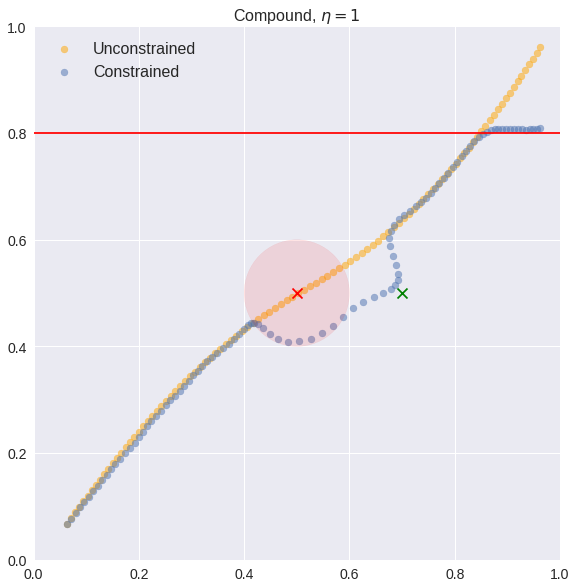}
    \includegraphics[width=0.3\textwidth]{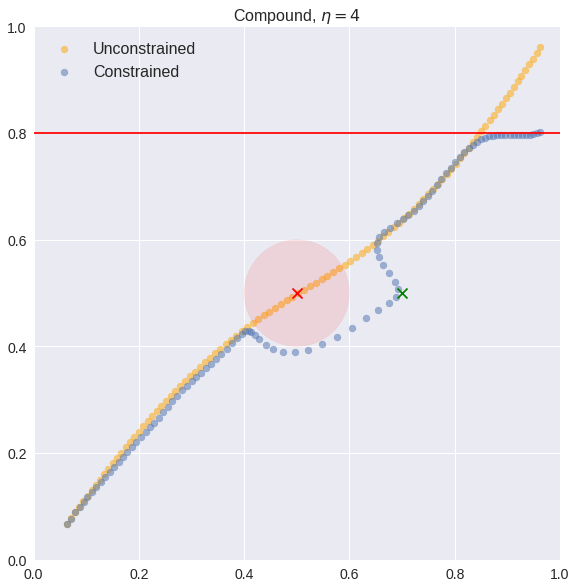}
    \caption{Avoid, Patrol and Until tests; and the Compound test, with $\eta=1$ and $\eta=4$}
    \label{fig:results}
\end{figure*}

Visually, we can confirm that the constrained training is actually satisfying the specifications of the Avoid, Patrol, and Compound (for $\eta=4$) tests in Fig.\ \ref{fig:results}. The Until and Compound (for $\eta=1$) tests are not fully satisfied (Until starts crossing the line $y=0.4$ slightly early, and Compound briefly enters the open ball and doesn't quite reach its goal at $o_2$), but clearly exhibit altered behaviour towards meeting the expected constraints. The slight violations happen because, in addition to the constraint loss, the trajectory loss plays a part in forming the neural network's training. These losses work against each other in our tests. Overall, though, these results demonstrate the clear effectiveness of the logical constraint in changing the learned behaviour of the DMP. 

Examining the Compound test (for $\eta=1$) more closely, we note that the trajectory slightly wanders within the open ball and does not quite reach the point we want it to. To verify how weighting affects this, we ran the test again with an increased value, namely $\eta=4$, applied to the constraint loss and the effects are obvious -- the trajectory now properly avoids the ball and gets closer to the point required. The same approach can be taken with the Until test to ensure satisfaction of its constraint. However, due to space considerations, we will not be considering this here.

At the conclusion of our experiments it is clear that the certainty of our approach comes at a cost in performance: running a full test can take anywhere from a few minutes to a few hours for the more complicated constraints, and the performance cannot be enhanced by running on a GPU as our OCaml functions cannot take advantage of this. There are several possible approaches to resolving this and improving performance as discussed next.

\section{Conclusion}
\label{sec:disc}

We demonstrate that using a theorem proving approach, we can formulate a deep embedding of \ltlf{} in higher-order logic and use this to fully formalise logical constraints for training, as well as the loss function and its derivative. We can then generate code for the whole logical framework and integrate it with PyTorch. Furthermore, our experimental results show that these constraints can successfully change the training process to match the desired behaviour. Thus, we believe that our work provides much stronger guarantees of correctness than one could expect from an ad-hoc implementation of logical operators made directly in Python.

There are some practical limitations to the current work. In particular, training the neural network can be slow because OCaml functions are used to compute the loss and its derivative with respect to each possible input, and these functions run outside the highly optimised PyTorch infrastructure. We are exploring some potential ways of addressing this shortcoming, which include adopting a tensor formulation within Isabelle to take advantage of OCaml bindings for PyTorch tensors \cite{ocamltensors} and producing PyTorch-compatible code by extending Isabelle's existing functionality for code generation.

Our approach is generic, so in principle a different formalism, e.g.\ a continuous-time logic, could be used instead of \ltlf{}.~Moreover, by formalising the derivative of the loss function, we unlock the potential to reason formally about the traversal of the loss surface during gradient descent. Given our results, we believe this work opens the way to a tighter integration between fully-formal symbolic reasoning in a theorem prover and machine learning.

\bibliography{constrainedtraining}

\newpage
\appendix

\section{Isabelle specifications}
\label{app:listings}

\begin{lstlisting}[basicstyle=\footnotesize\ttfamily, mathescape = true,caption=The {\ltlf{ }evaluation function, described in Section \ref{ssec:ltlf}},captionpos=b]{}
function eval :: "constraint $\Rightarrow$ path $\Rightarrow$ bool" where
  "eval c [] = False"
| "eval (Comp (Lequal v1 v2)) (s # ss) = ((s v1) $\leq$ (s v2))"
| "eval (Comp (Nequal v1 v2)) (s # ss) = ((s v1) $\neq$ (s v2))"
| "eval (Comp (Less v1 v2)) (s # ss) = ((s v1) < (s v2))"
| "eval (Comp (Equal v1 v2)) (s # ss) = ((s v1) = (s v2))"
| "eval (And c1 c2) (s # ss) = ((eval c1 (s # ss)) $\wedge$ (eval c2 (s # ss)))"
| "eval (Or c1 c2) (s # ss) = ((eval c1 (s # ss)) $\vee$ (eval c2 (s # ss)))"
| "eval (Next c) (s # ss) = eval c ss"
| "eval (Always c) (s # ss) = ((eval c (s # ss)) $\wedge$ (if ss = [] then True 
     else (eval (Always c) ss)))"
| "eval (Eventually c) (s # ss) = ((eval c (s # ss)) 
     $\vee$ (eval (Eventually c) ss))"
| "eval (Until c1 c2) (s # ss) = ((((eval c1 (s # ss)) 
     $\wedge$ (if ss = [] then True else (eval (Until c1 c2) ss)))) 
       $\vee$ eval c2 (s # ss))"
 | "eval (Release c1 c2) (s # ss) = (((((eval c2 (s # ss)) 
     $\wedge$ (eval (Release c1 c2) ss)))
       $\vee$ eval (And c1 c2) (s # ss)) $\wedge$ eval (Eventually c1) (s # ss))"
\end{lstlisting}

\begin{lstlisting}[basicstyle=\footnotesize\ttfamily, mathescape = true,caption={The soft gamma Min function and its correctness theorem, described in Section \ref{ssec:soft}},captionpos=b]{}
fun Min_gamma :: "real $\Rightarrow$ real $\Rightarrow$ real $\Rightarrow$ real" where
  "Min_gamma $\gamma$ a b = (if $\gamma$ $\leq$ 0 then min a b 
     else -$\gamma$ * ln (exp (-a/$\gamma$) + exp (-b/$\gamma$)))"

lemma Min_gamma_lim: "($\lambda$$\gamma$. Min_gamma $\gamma$ a b) $-$0$\rightarrow$ min a b"
\end{lstlisting}

\begin{lstlisting}[basicstyle=\footnotesize\ttfamily, mathescape = true,caption={The \loss{ }function, described in Section \ref{ssec:loss}},captionpos=b]{}
function L :: "constraint $\Rightarrow$ path $\Rightarrow$ real $\Rightarrow$ real" where
  "L c [] $\gamma$ = 1"
| "L (Comp (Lequal v1 v2)) (s # ss) $\gamma$ = Max_gamma $\gamma$ (s v1 - s v2) 0"
| "L (Comp (Nequal v1 v2)) (s # ss) $\gamma$ = Nequal_gamma $\gamma$ (s v1) (s v2)"
| "L (Comp (Less v1 v2)) (s # ss) $\gamma$ = Max_gamma $\gamma$ 
     (L (Comp (Lequal v1 v2)) (s # ss) $\gamma$) 
     (L (Comp (Nequal v1 v2)) (s # ss) $\gamma$)"
| "L (Comp (Equal v1 v2)) (s # ss) $\gamma$ = Max_gamma $\gamma$ 
     (L (Comp (Lequal v1 v2)) (s # ss) $\gamma$) 
     (L (Comp (Lequal v2 v1)) (s # ss) $\gamma$)"
| "L (And c1 c2) (s # ss) $\gamma$ = Max_gamma $\gamma$ 
     (L c1 (s # ss) $\gamma$) 
     (L c2 (s # ss) $\gamma$)"
| "L (Or c1 c2) (s # ss) $\gamma$ = Min_gamma $\gamma$ 
     (L c1 (s # ss) $\gamma$) 
     (L c2 (s # ss) $\gamma$)"
| "L (Next c) (s # ss) $\gamma$ = L c ss $\gamma$"
| "L (Always c) (s # ss) $\gamma$ = Max_gamma $\gamma$ (L c (s # ss) $\gamma$) 
     (if ss = [] then 0 else (L (Always c) ss) $\gamma$)"
| "L (Eventually c) (s # ss) $\gamma$ = Min_gamma $\gamma$ (L c (s # ss) $\gamma$) 
     (L (Eventually c) ss $\gamma$)"
| "L (Until c1 c2) (s # ss) $\gamma$ = Min_gamma $\gamma$ (L c2 (s # ss) $\gamma$) 
     (Max_gamma $\gamma$ (L c1 (s # ss) $\gamma$) (if ss = [] then 0 
       else (L (Until c1 c2) ss $\gamma$)))"
| "L (Release c1 c2) (s # ss) $\gamma$ = Max_gamma $\gamma$ 
     (L (Eventually c1) (s # ss) $\gamma$) 
     (Min_gamma $\gamma$
       (Max_gamma $\gamma$ (L c1 (s # ss) $\gamma$) (L c2 (s # ss) $\gamma$))
       (Max_gamma $\gamma$ (L c2 (s # ss) $\gamma$) (L (Release c1 c2) ss $\gamma$)))"
\end{lstlisting}

\begin{lstlisting}[basicstyle=\footnotesize\ttfamily, mathescape = true,caption={The \dloss{ }function, described in Section \ref{sec:dloss}},captionpos=b]{}
function dL :: "constraint $\Rightarrow$ path $\Rightarrow$ real $\Rightarrow$ int $\Rightarrow$ int $\Rightarrow$ real" where
  "dL c [] $\gamma$ i j = 0"
| "dL (Comp (Lequal v1 v2)) (s # ss) $\gamma$ i j = (if i $\neq$ 0 
     then 0 else (dLequal_gamma_ds $\gamma$ (s v1) 
       (if eqint j v1 then (1::real) else 0) (s v2) 
       (if eqint j v2 then (1::real) else 0)))"
| "dL (Comp (Nequal v1 v2)) (s # ss) $\gamma$ i j = (if (i $\neq$ 0) then 0 
     else (dNequal_gamma_ds $\gamma$ (s v1) 
       (if j = v1 then 1 else 0) (s v2) 
       (if j = v2 then 1 else 0)))"
| "dL (Comp (Less v1 v2)) (s # ss) $\gamma$ i j = dMax_gamma_ds $\gamma$
  (L (Comp (Lequal v1 v2)) (s # ss) $\gamma$)
    (dL (Comp (Lequal v1 v2)) (s # ss) $\gamma$ i j)
  (L (Comp (Nequal v1 v2)) (s # ss) $\gamma$)
    (dL (Comp (Nequal v1 v2)) (s # ss) $\gamma$ i j)"
| "dL (Comp (Equal v1 v2)) (s # ss) $\gamma$ i j = dMax_gamma_ds $\gamma$
  (L (Comp (Lequal v1 v2)) (s # ss) $\gamma$)
    (dL (Comp (Lequal v1 v2)) (s # ss) $\gamma$ i j)
  (L (Comp (Lequal v2 v1)) (s # ss) $\gamma$)
    (dL (Comp (Lequal v2 v1)) (s # ss) $\gamma$ i j)"
| "dL (And c1 c2) (s # ss) $\gamma$ i j = dMax_gamma_ds $\gamma$
  (L c1 (s # ss) $\gamma$)
    (dL c1 (s # ss) $\gamma$ i j)
  (L c2 (s # ss) $\gamma$)
    (dL c2 (s # ss) $\gamma$ i j)"
| "dL (Or c1 c2) (s # ss) $\gamma$ i j = dMin_gamma_ds $\gamma$
  (L c1 (s # ss) $\gamma$)
    (dL c1 (s # ss) $\gamma$ i j)  
  (L c2 (s # ss) $\gamma$)
    (dL c2 (s # ss) $\gamma$ i j)"
| "dL (Next c) (s # ss) $\gamma$ i j = dL c ss $\gamma$ (i-1) j"
| "dL (Always c) (s # ss) $\gamma$ i j = dMax_gamma_ds $\gamma$
  (L c (s # ss) $\gamma$)
    (dL c (s # ss) $\gamma$ i j)
  (if ss = [] then 0 else (L (Always c) ss) $\gamma$)
    (if ss = [] then 0 else (dL (Always c) ss) $\gamma$ (i-1) j)"
| "dL (Eventually c) (s # ss) $\gamma$ i j = dMin_gamma_ds $\gamma$
  (L c (s # ss) $\gamma$)
    (dL c (s # ss) $\gamma$ i j)
  (L (Eventually c) ss $\gamma$)
    (dL (Eventually c) ss $\gamma$ (i-1) j)"
| "dL (Until c1 c2) (s # ss) $\gamma$ i j = dMin_gamma_ds $\gamma$ (L c2 (s # ss) $\gamma$)
     (dL c2 (s # ss) $\gamma$ i j) (Max_gamma $\gamma$ (L c1 (s # ss) $\gamma$) 
     (if (ss = []) then 0 else (L (Until c1 c2) ss $\gamma$)))
     (dMax_gamma_ds $\gamma$ (L c1 (s # ss) $\gamma$) (dL c1 (s # ss) $\gamma$ i j)
       (if (ss = []) then 0 else (L (Until c1 c2) ss $\gamma$))
       (if (ss = []) then 0 else (dL (Until c1 c2) ss $\gamma$ (i-1) j)))"
| "dL (Release c1 c2) (s # ss) $\gamma$ i j = dMax_gamma_ds $\gamma$
  (L (Eventually c1) (s # ss) $\gamma$) 
    (dL (Eventually c1) (s # ss) $\gamma$ i j)
  (Min_gamma $\gamma$ (Max_gamma $\gamma$ (L c1 (s # ss) $\gamma$) (L c2 (s # ss) $\gamma$))
  (Max_gamma $\gamma$ (L c2 (s # ss) $\gamma$) (L (Release c1 c2) ss $\gamma$)))
    (dMin_gamma_ds $\gamma$ (Max_gamma $\gamma$ (L c1 (s # ss) $\gamma$) (L c2 (s # ss) $\gamma$))
      (dMax_gamma_ds $\gamma$
      (L c1 (s # ss) $\gamma$) (dL c1 (s # ss) $\gamma$ i j)
      (L c2 (s # ss) $\gamma$) (dL c2 (s # ss) $\gamma$ i j))
    (Max_gamma $\gamma$ (L c2 (s # ss) $\gamma$) (L (Release c1 c2) ss $\gamma$))
      (dMax_gamma_ds $\gamma$
      (L c2 (s # ss) $\gamma$) (dL c2 (s # ss) $\gamma$ i j)
      (L (Release c1 c2) ss $\gamma$) (dL (Release c1 c2) ss $\gamma$ (i-1) j)))"
\end{lstlisting}

\section{A brief introduction to \ltlf{}}
\label{app:ltlf}

Linear temporal logic uses time-based modalities with which propositions can be held to be true or false at particular steps of a sequence in discrete time. This truth-value may change for a given proposition as the time-sequence progresses.

The \ltlf{} formulae are: $\rho$, $\rho_1 \wedge \rho_2$, $\rho_1 \vee \rho_2$, $\mathcal{N}\rho$ (Next), $\square \rho$ (Always), $\Diamond \rho$ (Eventually), $\rho_1 \mathcal{U} \rho_2$ (Weak Until), $\rho_1 \mathcal{R} \rho_2$ (Strong Release). The first three are understood as per the semantics of propositional logic connectives. Informally, the rest are to be understood as follows:
\begin{itemize}
	\item $\mathcal{N}\rho$ (Next $\rho$) is true if at the \textit{next} step along the time-sequence, $\rho$ holds. Note that as we are working in finite paths, we must consider how to evaluate this at the termination of a path -- we discuss this below.
	\item $\square\rho$ (Always $\rho$) holds if at the current step and all subsequent steps along the time-sequence, $\rho$ holds.
	\item $\Diamond\rho$ (Eventually $\rho$) holds if at the current or at least one subsequent step along the time-sequence, $\rho$ holds.
	\item $\rho_1\mathcal{U}\rho_2$ ($\rho_1$ \texttt{Until} $\rho_2$) means that $\rho_1$ holds at least for all steps until $\rho_2$ holds. $\rho_2$ need not hold at any future point -- this is a Weak \texttt{Until}.
	\item $\rho_1\mathcal{R}\rho_2$ ($\rho_1$ \texttt{Release} $\rho_2$) means that $\rho_2$ holds at least until and including the point when $\rho_1$ holds. $\rho_1$ must hold at some point in the path -- this is a Strong \texttt{Release}.
\end{itemize}
As discussed in Section \ref{ssec:ltlf}, we must consider how $\mathcal{N}$ is intended to behave at the end of a time-sequence. As per the usual treatment of LTL on a finite trace, that with a sequence of length $i$, we assume $\neg(\mathcal{N} \rho)$ holds at the final step $i$ \cite{de2013linear}.

\section{An extended description of soft functions}
\label{app:soft}

The soft maximum function has a derivative function that we briefly mentioned in Section \ref{ssec:soft}. We give some more details of its construction here.

\begin{lstlisting}[basicstyle=\footnotesize\ttfamily, mathescape = true,caption={The derivative for our soft max function, as discussed in Section \ref{ssec:soft}},captionpos=b]{}
fun dMax_gamma_da :: "real $\Rightarrow$ real $\Rightarrow$ real $\Rightarrow$ real" where
  "dMax_gamma_da $\gamma$ a b = exp(a/$\gamma$)/(exp (a/$\gamma$) + exp (b/$\gamma$))"

fun dMax_gamma_ds :: "real $\Rightarrow$ real $\Rightarrow$ real $\Rightarrow$ real $\Rightarrow$ real $\Rightarrow$ real" where
  "dMax_gamma_ds $\gamma$ a da b db = (dMax_gamma_da $\gamma$ a b) * da 
     + (dMax_gamma_db $\gamma$ a b) * db"
\end{lstlisting}

For \texttt{Max\_gamma}, the derivatives with respect to $a$ and $b$ are built separately before being combined to give the \texttt{dMax\_gamma\_ds} function.~In the above, \texttt{dMax\_gamma\_da} is defined and then proven to be the partial derivative of the softmax function \texttt{Max\_gamma} $\gamma$ $a$ $b$ w.r.t.~$a$ when $\gamma > 0$. The formulation of the derivative with respect to $b$ is similar and hence omitted. We only consider the case $\gamma > 0$ as this is when our soft functions are differentiable. \texttt{dMax\_gamma\_ds} is the derivative w.r.t.~both paramaters.

In addition to the soft maximum and minimum functions we discuss in Section \ref{ssec:soft}, there are a number of other soft functions we use. We discuss their Isabelle specification and their proofs of correctness here. 

We use a function \texttt{Bell\_gamma} to capture losses from the \texttt{Nequal} comparison. In Isabelle, we have:

\begin{lstlisting}[basicstyle=\footnotesize\ttfamily, mathescape = true,caption={The \texttt{Nequal\_gamma} function},captionpos=b]{}
fun Nzero :: "real $\Rightarrow$ real" where
  "Nzero x = (if x=0 then 1 else 0)"

fun Bell_gamma :: "real $\Rightarrow$ real $\Rightarrow$ real" where
  "Bell_gamma $\gamma$ x = (if $\gamma$$\leq$0 then (Nzero x) else 1/exp(x^2/(2*$\gamma$)))"
  
fun Nequal_gamma :: "real $\Rightarrow$ real $\Rightarrow$ real $\Rightarrow$ real" where 
  "Nequal_gamma $\gamma$ a b = Bell_gamma $\gamma$ (a-b)"
\end{lstlisting}

\noindent where we have a derivative function, \texttt{dBell\_gamma\_dx} defined as:

\begin{lstlisting}[basicstyle=\footnotesize\ttfamily, mathescape = true,caption={Derivative for the \texttt{Bell\_gamma} function},captionpos=b]{}
fun dBell_gamma_dx :: "real $\Rightarrow$ real $\Rightarrow$ real" where
  "dBell_gamma_dx $\gamma$ x = exp (-(x^2)/(2*$\gamma$))*(-x/$\gamma$)"
\end{lstlisting}

We also prove that \texttt{lim$_{\gamma\to0}$\,Bell\_gamma $\gamma$ 0 = 1} (and is 0 for any other value) and that \texttt{dBell\_gamma\_dx} is the expected derivative when $\gamma>0$:
\begin{lstlisting}[basicstyle=\footnotesize\ttfamily, mathescape = true,caption={Correctness of the \texttt{Bell\_gamma} derivative},captionpos=b]{}
theorem Bell_gamma_deriv_gamma_gt_zero:
  assumes gamma_gt_zero: "$\gamma$>0"
  shows "(($\lambda$y. Bell_gamma $\gamma$ y) has_field_derivative 
    dBell_gamma_dx $\gamma$ x) (at x)"
\end{lstlisting}

As can be seen above, \texttt{Nequal\_gamma} is defined by passing the difference of two values $a$ and $b$ to \texttt{Bell\_gamma}. We can thus leverage the above properties to define and prove its derivative \texttt{dNequal\_gamma\_ds}, in a similar way as was done for \texttt{Max\_gamma}. For brevity, this is omitted.

\section{PyTorch integration}
\label{app:pytorchint}

PyTorch is a Python library for deep learning \cite{paszke2019pytorch}. We use it for testing whether our generated code enables the intended training within the framework developed by Innes and Ramamoorthy \cite{innes2020elaborating_rss} (see Section \ref{sec:exper}). Crucial to PyTorch (and other similar libraries) is the tensor datatype that is used to represent inputs and outputs over a network of mathematical operations in a typically multi-dimensional array-like form.

A sequence of tensors and operations performed thereon are recorded in a directed acyclic graph known as a computational graph. Traversing this graph from a root scalar tensor, PyTorch's automatic differentiation engine \texttt{torch.autograd} computes the gradients of the root tensor with respect to each of the elements of the tensors in the graph, via successive application of the chain rule of differentiation. This algorithm enables a fundamental approach to neural network training, backpropagation, where a network's weights are iteratively adjusted according to the gradient of a computed loss with respect to themselves.

In order for our LTL loss function to form part of a computational graph in PyTorch, it must be implemented as a subclass of \texttt{autograd.Function}. We call this class \texttt{LTL\_Loss} as per our OCaml module, and it is parameterised by an LTL constraint, represented as an OCaml expression, constants for comparison, and a value for $\gamma$.

As noted in Section \ref{ssec:statepath}, an instantiation of a path can be represented by a matrix:

\[
P = \begin{bmatrix}
    \mathbf{s}_0 \\ \vdots \\ \mathbf{s}_{N-1}
    \end{bmatrix} = \begin{bmatrix} 
    t_{00} & \dots & t_{0(K-1)}\\
    \vdots & \ddots & \vdots\\
    t_{(N-1)0} & \dots & t_{(N-1)(K-1)} 
    \end{bmatrix}
\]
Here, each \textit{row} is a representation of a state along the path. This matrix is useful as it can be implemented as a PyTorch tensor.

To interface our OCaml code for \loss{} and \dloss{} with \texttt{autograd}, two methods are defined:

\begin{enumerate}
    \item \texttt{forward}: this applies the loss function to a 2-dimensional input tensor representing a path, using the OCaml bindings to apply the OCaml representation of $\mathcal{L}$. If a 3-dimensional input tensor is given, with the leading dimension indexing separate paths, an average loss is computed.
    \item \texttt{backward}: this computes the derivative of the scalar loss output with respect to the input tensor. Iterating over each element of the tensor, we call the OCaml representation of $d\mathcal{L}$ to compute the derivative of the loss with respect to it.
\end{enumerate}

With this, our class is functionally identical to a differentiable PyTorch operation on tensors,  as discussed in Section \ref{ssec:pytorchint}.

\end{document}